# Obtaining Depth Maps From Color Images By Region Based Stereo Matching Algorithms

Barış Baykant ALAGÖZ

***Abstract:*** *In the paper, region based stereo matching algorithms are developed for extraction depth information from two color stereo image pair. A filter eliminating unreliable disparity estimation was used for increasing reliability of the disparity map. Obtained results by algorithms were presented and compared.*

## Introduction:

Obtaining reliable depth maps, indicating distance of surface from the stereo camera pair, have importance in robotic applications and autonomous systems. Intelligent systems, which can move around by itself, could be developed by obtaining dept information from the sensors. Stereovision is the one of methods that can yield dept information of the scene. It uses stereo image pairs from two cameras to produce disparity maps that can be easily turn into dept maps. Reliability of depth maps and computational cost of algorithm is key issue for implementing real time robust applications.

Many studies suggesting reliable algorithms were made [1-3] and some studies comparing performance of the algorithm can be found in literature.[4] In this study, we will develop two different algorithms based on region growing for color images and compare their results in the aspect of reliability and complexity.

## Region Based Stereo Algorithms:

a) Global Error Energy Minimization by Smoothing Functions

In this method, we used block-matching technique in order to construct an Error Energy matrix for every disparity. Lets denote left image in RGB format by $L(i,j,c)$, denote right image in RGB format by $R(i,j,c)$ and error energy by $e(i,j,d)$. For $n \times m$ window size of block matching, error energy $e(i,j,d)$ can be expressed by,

$$e(i,j,d) = \frac{1}{3 \cdot n \cdot m} \cdot \sum_{x=i}^{i+n} \sum_{y=j}^{j+m} \sum_{k=1}^{3} (L(x, y+d, k) - R(x, y, k))^2 \qquad (1)$$

where, c represents RGB components of images and takes value of $\{1,2,3\}$ corresponding to red, blue and green. $d$ is the disparity. For a predetermined disparity search range ($w$), every $e(i,j,d)$ matrix respect to disparity is smoothed by applying averaging filter many times. (See Figure 1.b) Averaging filter (linear filter)[5] removes very sharp change in energy which possibly belong to incorrect matching. An other important properties of repeating application of averaging filter is that it makes apparent global trends in energy. (Local filtering in iterations could solve a global total variational optimization problem) [6] Considering global trend in error energy naturally makes this algorithm a region based algorithm. For $n \times m$ window size, averaging filtering of $e(i,j,d)$ can be expressed by following equation,

$$\tilde{e}(i,j,d) = \frac{1}{n \cdot m} \sum_{x=i}^{i+n} \sum_{y=j}^{j+m} e(x,y,d) \qquad (2)$$

After iterative application of averaging filtering to error energy for each disparity, we selected the disparity ($d$), which has minimum error energy $\tilde{e}(i,j,d)$ as the most reliable disparity estimation for pixel $(i,j)$ of disparity map. (See Figure 1.c) Lets write basic steps of algorithm more properly,

**Step 1**: For every disparity $d$ in disparity search range, calculate error energy matrix. (Figure 1.a)
**Step 2**: Apply average filtering iteratively to every error matrix calculated for a disparity value in the range of disparity search range. (Figure 1.b)
**Step 3**: For every $(i,j)$ pixel, find the minimum error energy $\tilde{e}(i,j,d)$, assign its disparity index ($d$) to $d(i,j)$ which is called disparity map. (Figure 1.c)

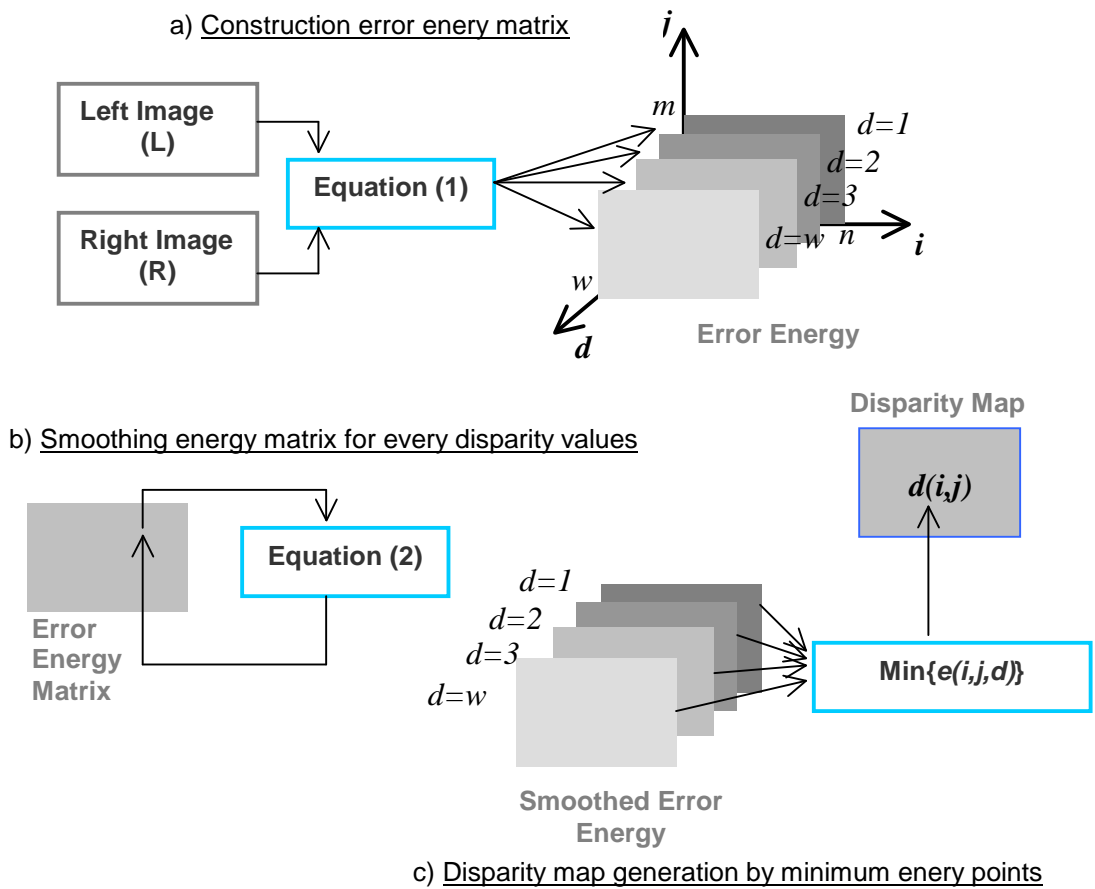

**Figure 1.** Method using global error energy minimization by smoothing functions

b) Line Growing Based Stereo Matching

In this section, we proposed an algorithm based on region growing. In this manner, we consider region-growing mechanism in two phases operation. First phase, finding root point to grow region (*Root Selection process*) and the second phase, growing region for a root point corresponding to predefined rule.(*Region Growing process*) Our rule for associating a point to root point in the growing process is to have lower error energy than a predetermined threshold of error energy (*LineGrowingThreshold*). In our application, being associated to a root points means to have the same disparity by root point. Thereby, the region emerged from all associated points has a disparity value. Actually, we should call the algorithm as disparity growing. Lets generally express steps of the algorithm in a list,

**Step 1**: (*Root Selection process*) Select a point, which isn't belonging to any grown region and find its disparity using energy function equation (1). Set it root point and set its disparity to region disparity then go to step 2. If you didn't find any disparity with lower enough error energy, repeat this step for the next point.
**Step 2**: (*Region Growing process*) Calculates error energy of neighbor points just for root point disparity, which was called region disparity. If it is lower than the predetermined error energy threshold, associate this point to region. Otherwise, left it free.
**Step 3**: Proceed the Step 2 until region growing any more. In the case that region growing is completed, turn back to step 1 to find out new root point to repeat these steps. When all points in image processed, stop the algorithm. Grown disparity regions composes disparity map $d(i, j)$.

In order to reduce complexity of the algorithm, we allow the region growing in the direction of rows since disparity of stereo image is only in row directions. So, only one neighbor, which is the point after searched point is inspected for region growing. (See Figure 2) We called this type of algorithm as line growing. Lets revise the algorithm steps for line growing,

**Step 1**: (*Root Selection process*) Search on the rows to find a root point, which isn't belonging to any grown region and then find its disparity using energy function by line type window. If the error energy of selected point is equal or lower than *LineGrowingThreshold* ($V_{LG}$), select it as root point and go to step 2. If not, marked the point idle and do Step1 for following point in the row.
**Step 2**: (*Region Growing process*) Calculate error energy of next neighbor point by root point disparity. If it is equal or lower than the predetermined error energy threshold $V_{LG}$, associate this point to region. Otherwise, back to step 1 to find a new root point.
**Step 3**: Proceed the Step 1 and Step 2 row by row until reaching end point of image. Grown disparity regions compose of the disparity map $d(i, j)$.

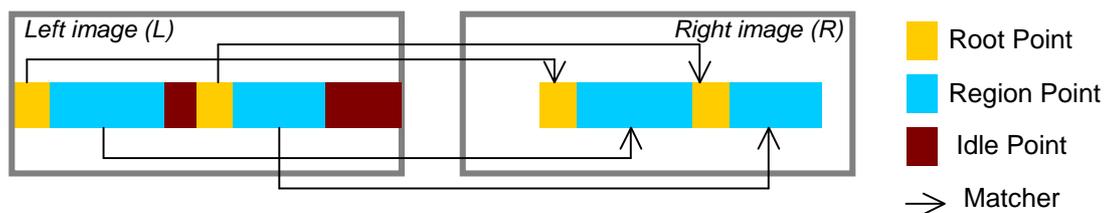

**Figure 2.** Method using line growing

## Depth Map Generation From Disparity Map:

To better understand depth and disparity relation, let see stereo projection representation illustrated in the Figure 3. By considering the figure, one can derive relation between dept ($Z$) and disparity ($d$) by using basic geometrical calculations as following,

$$Z(i,j) = f \cdot \frac{T}{d(i,j)} \qquad (4)$$

If real location of object surface projected at pixel $(i,j)$ is willing to calculate, following formulas can be used in calculation of $(X,Y)$ points after calculation of the $Z$.

$$X = \frac{(Z-f)}{f} \cdot i \text{ and } Y = \frac{(Z-f)}{f} \cdot j \qquad (5)$$

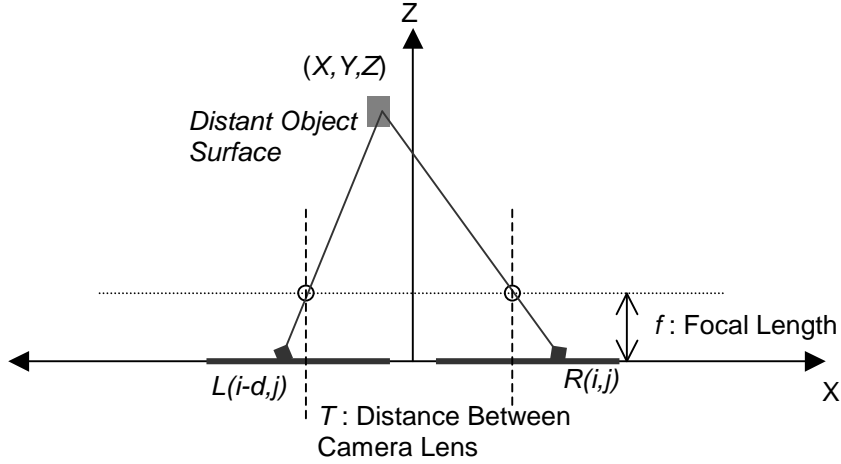

**Figure 3.** Representation of the stereo projection

In order to obtain smoother depth map to be used in applications such as robot navigation, $5x5$ window sized median filtering should be applied to disparity ($d$) before computing dept ($Z$).

## Filtering Unreliable Disparity Estimation By Average Error Thresholding Mechanism:

We define reliability ($R_d$) of the obtained disparity map $d$ by mean value of the error energy of disparity maps ($E_d$). It can be expressed as,

$$R_d = \frac{1}{Mean(E_d - \{ne\})} = \frac{1}{S_d} \cdot \left( \sum_{(i,j)-\{E_d(i,j)=ne\}}^{(n,m)} E_d(i,j) \right)^{-1} \qquad (5)$$

Error energy of disparity maps ($E_d$) can be written as following,

$$E_d(i,j) = \frac{1}{3 \cdot n \cdot m} \cdot \sum_{x=i}^{i+n} \sum_{y=j}^{j+m} \sum_{k=1}^{3} (L(x, y+d(i,j), k) - R(x, y, k))^2 \qquad (6)$$

Disparity map contains some unreliable disparity estimations for some points around the object boundaries mostly as a result of object occultation in images. These unreliable disparities can be detected by observing high error energy in the $E_d$. In order to increase reliability of obtained disparity map $d(i,j)$, simple thresholding mechanism , described by equation (7), can be applied to filter some unreliable disparity estimations in the $d(i,j)$.

$$\tilde{d}(i,j) = \begin{cases} d(i,j) & E_d(i,j) \leq Ve \\ 0 & E_d(i,j) > Ve \end{cases} \qquad (7)$$

$$\tilde{E}_d(i,j) = \begin{cases} E_d(i,j) & E_d(i,j) \leq Ve \\ ne & E_d(i,j) > Ve \end{cases} \qquad (8)$$

$\tilde{d}(i,j)$ will be the more reliable version of $d(i,j)$ by filtering some unreliable disparity estimations. Setting disparity to $ne$ in equation (8) refers "no-estimated" state and $E_d(i,j)$ values that have $ne$ state is excluded in calculation of $R_d$. $S_d$ parameter in the equation (5) represents the number of points in error energy which are not $ne$. $\tilde{E}_d$ is error energy for $\tilde{d}(i,j)$. $Ve$ is error energy threshold for deciding disparity estimation to be unreliable one. To determining $Ve$ automatically, we used following formula.

$$Ve = \alpha \cdot Mean(E_d) \qquad (9)$$

In the equation (6), $\alpha$ is tolerance coefficient to adjust reliability of the filtering process. Decreasing the $\alpha$ leads the $\tilde{d}$ more reliable. (See proof of the statement in the section of *Propositions*) Unfortunately, decreasing the $\alpha$ will erode disparity map because of eliminating more disparity points in the map.

## Obtained Results:

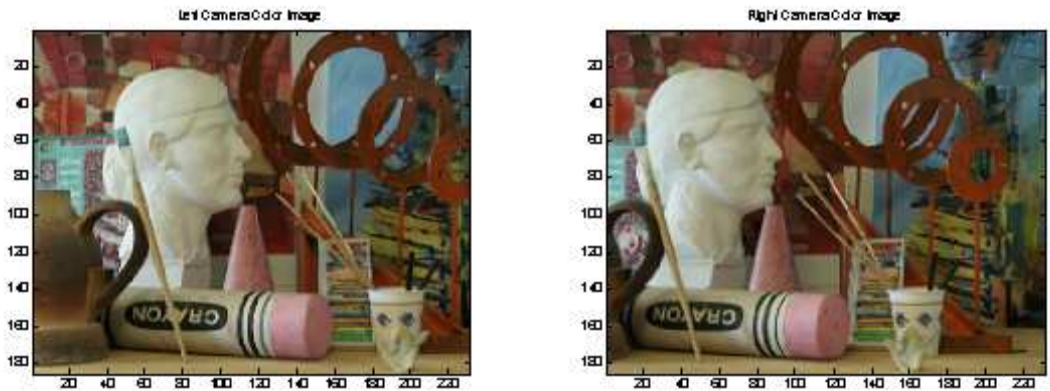

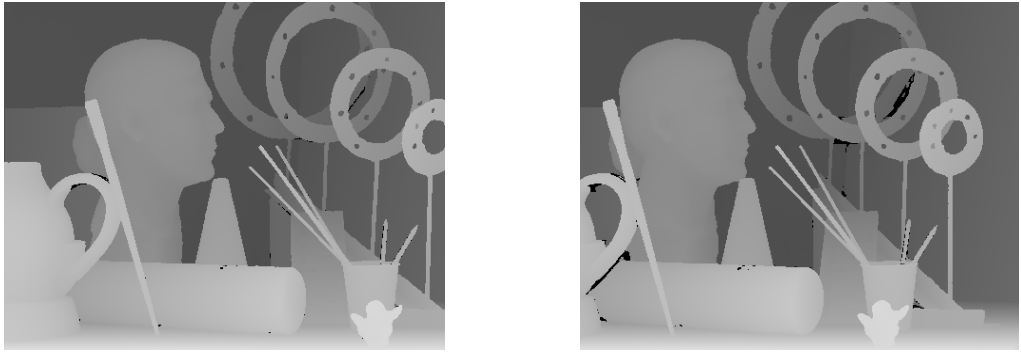

**Figure 4.** Color stereo image pair and their ground truth images. ( from vision.middlebury.edu/stereo/ )

Test images and their ground-truth images are seen in the Figure 4. Results of algorithms introduced in the paper for the test images were given in the following sub-sections.

a) Results of Global Error Energy Minimization by Smoothing Functions

*__Results for point matching window (1x1)__*:
[ $n=1, m=1, d_{max}=40, f=30, T=20, \alpha=1$ ]

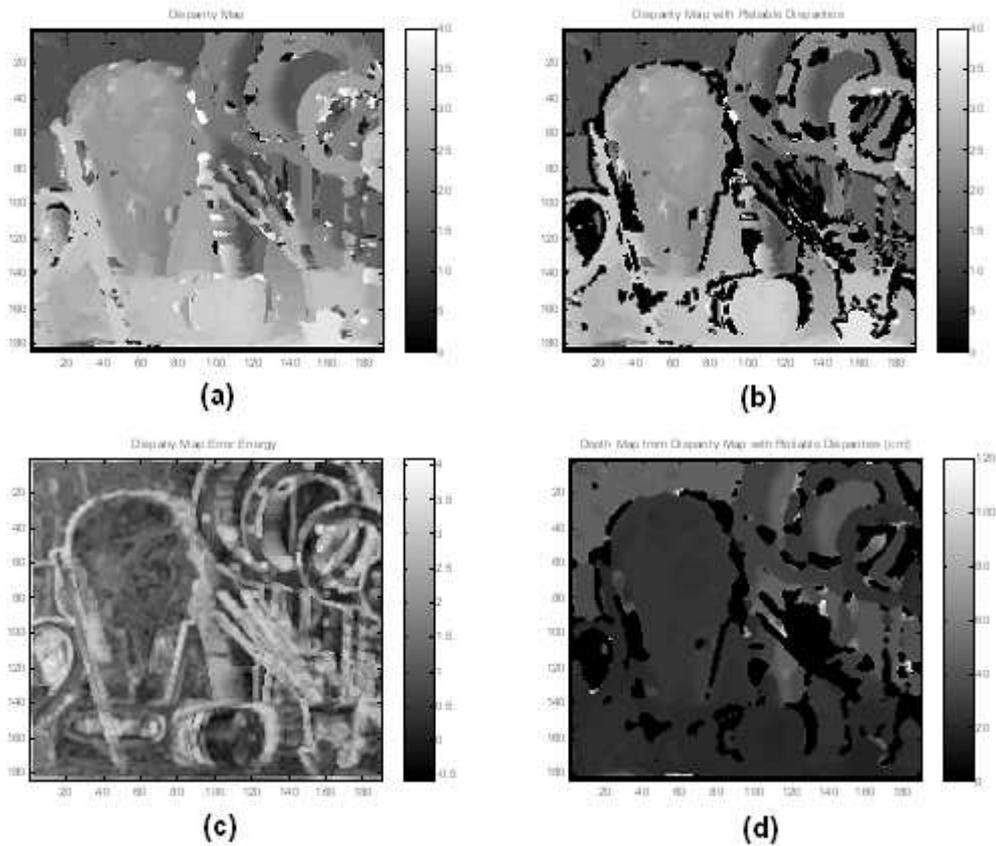

**Figure 5.** Results obtained from point matching window. (a) $d$, (b) $\tilde{d}$, (c) $E_d$, (f) $Z$

***Results for line matching window (1x5)***:
[$n=1, m=5, d_{max}=40, f=30, T=20, \alpha=1$]

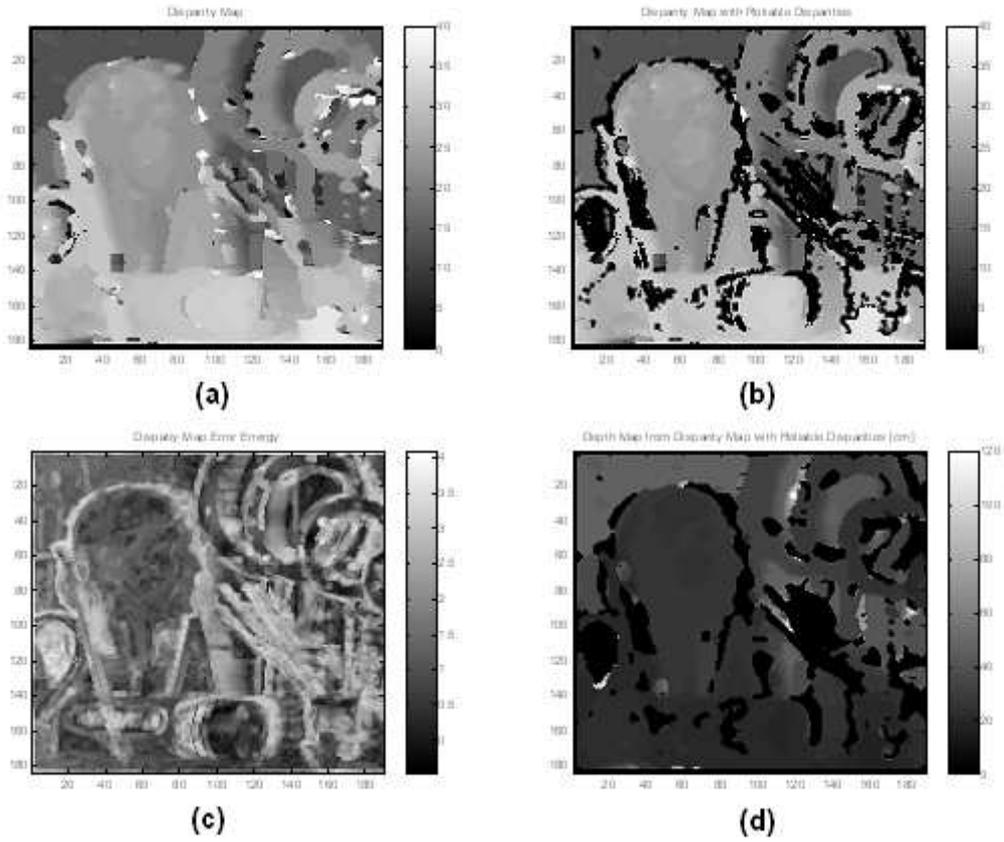

**Figure 6.** Results obtained from line matching window. (a) $d$, (b) $\tilde{d}$, (c) $E_d$, (f) $Z$

***Results for 3x3 matching window (3x3)***:
[$n=3, m=3, d_{max}=40, f=30, T=20, \alpha=1,$]

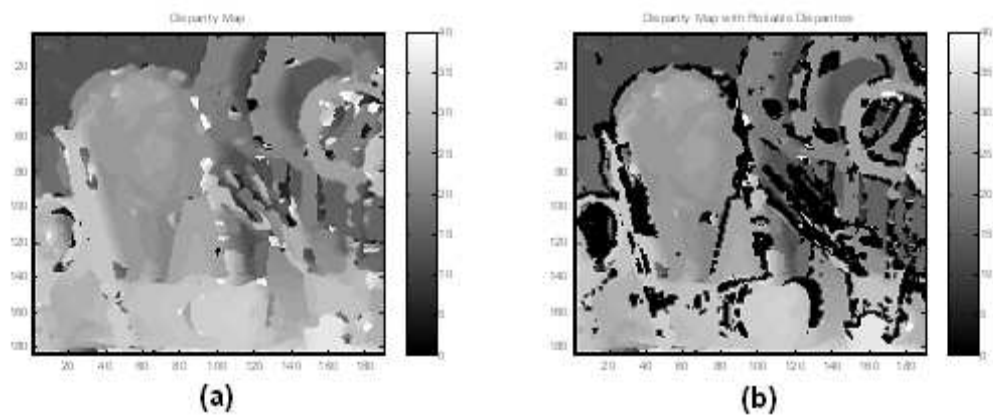

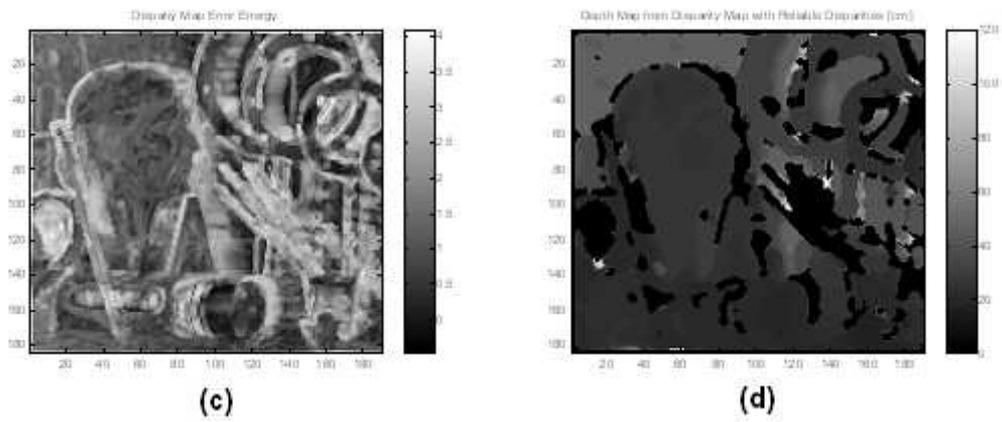

**Figure 7.** Results obtained from region matching window. (a) $d$, (b) $\tilde{d}$, (c) $E_d$, (f) $Z$

b) Results of Line Growing

*Results for Line Growing with Threshold of 60*:
[$n=1, m=5$, $V_{LG}=60$, $d_{max}=40$, $f=30, T=20, \alpha=1$]

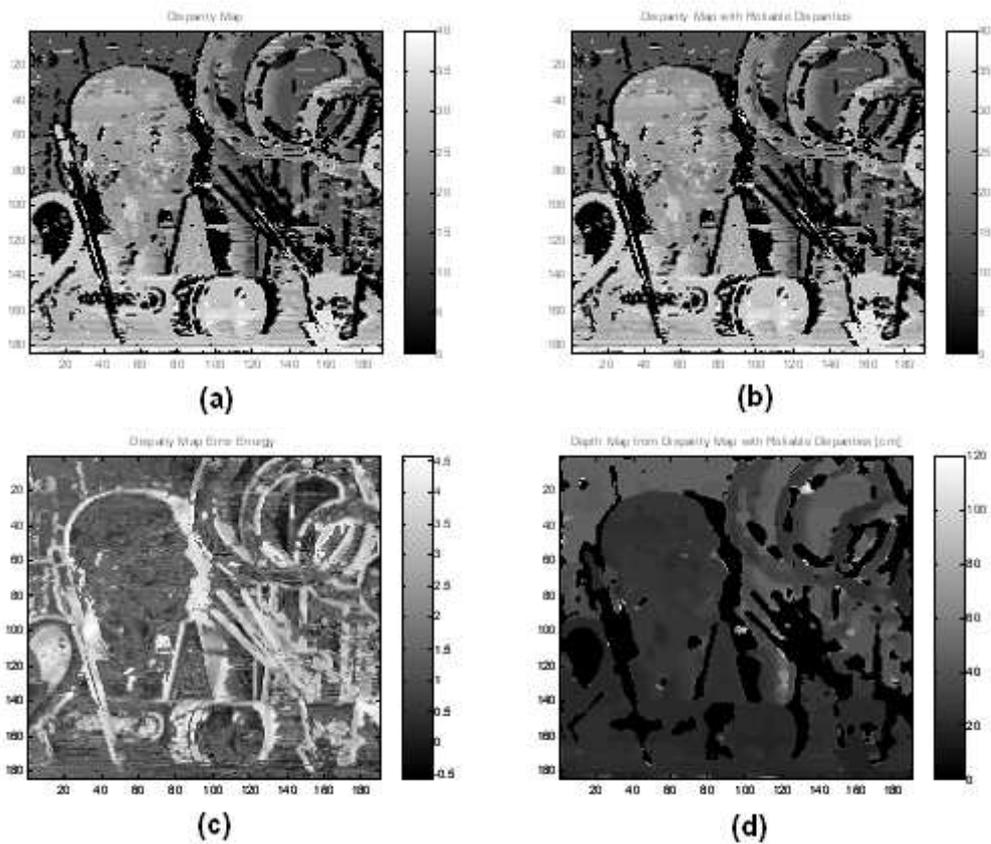

**Figure 8.** Results obtained for $V_{LG}=60$. (a) $d$, (b) $\tilde{d}$, (c) $E_d$, (f) $Z$

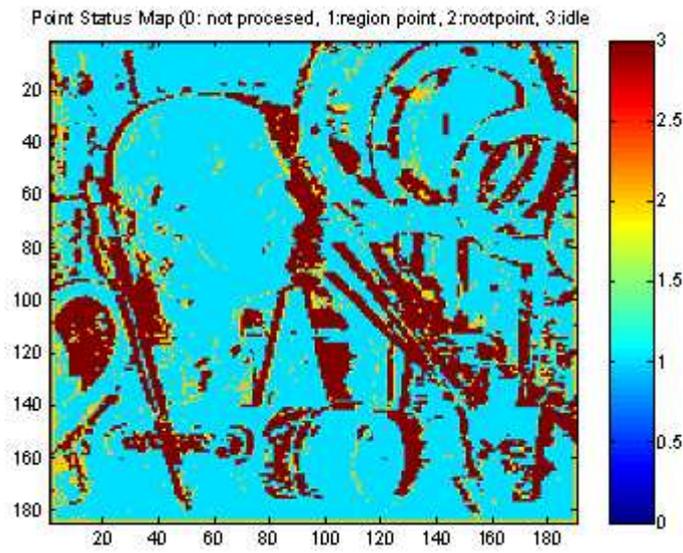

**Figure 9.** Point status Map for $V_{LG} = 60$.
[0:not processed, 1:region point, 2:root point, 3:idle point]

***Results for Line Growing with Threshold of 10***:
[ $n = 1, m = 5$, $V_{LG} = 10$, $d_{max} = 40$, $f = 30, T = 20$, $\alpha = 1$ ]

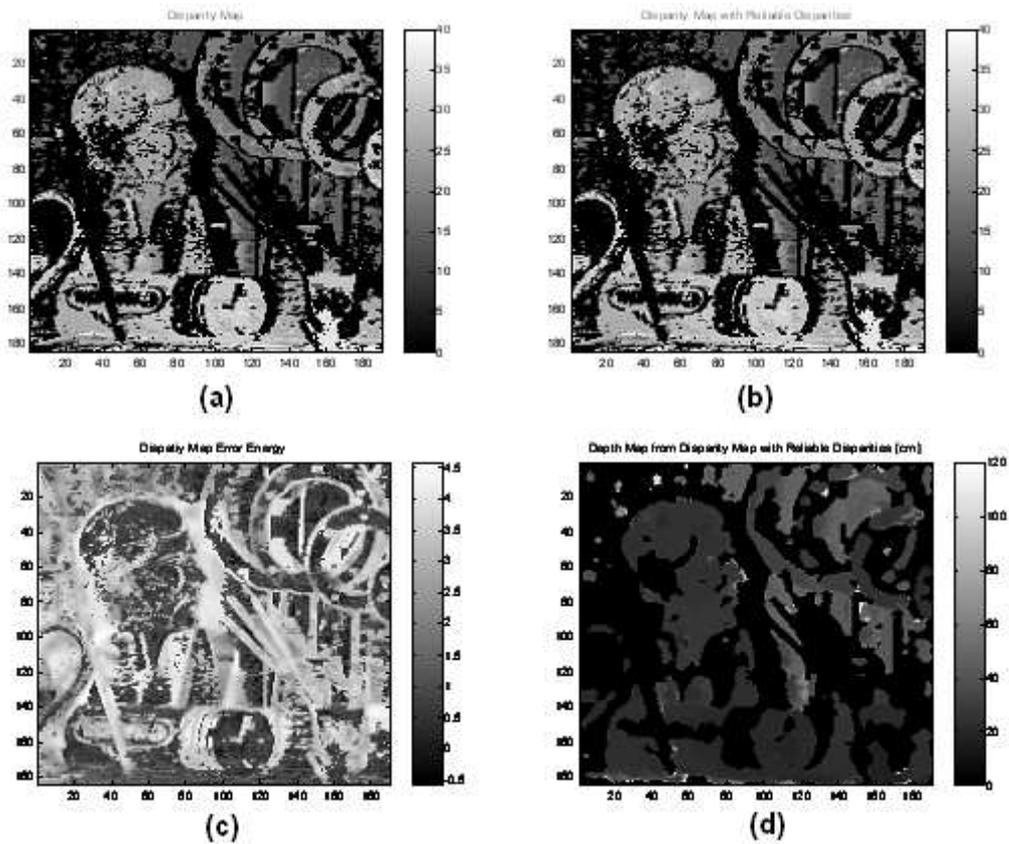

**Figure 10.** Results obtained for $V_{LG} = 10$. (a) $d$, (b) $\tilde{d}$, (c) $E_d$, (f) $Z$

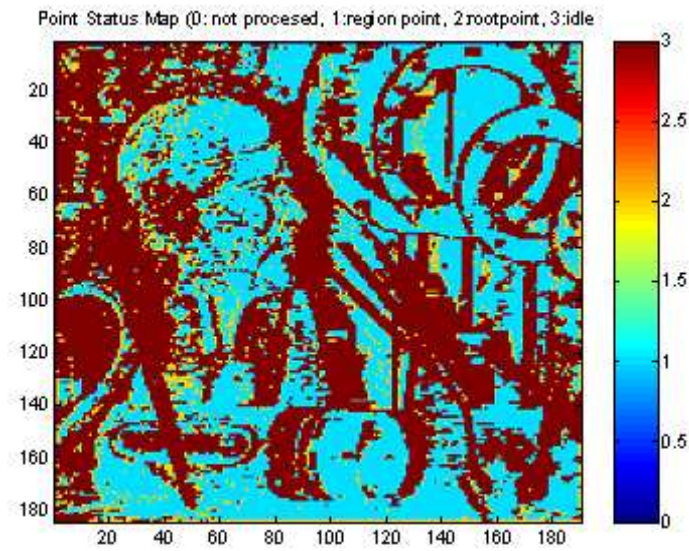

**Figure 11.** Point status Map for $V_{LG} = 10$.
[0:not processed, 1:region point, 2:root point, 3:idle point]

c) Reliability And Speed Comparisons

Reliability comparison for the results illustrated in the Figure 5, 6, 7, 8 and 10 is given in the Figure 12.

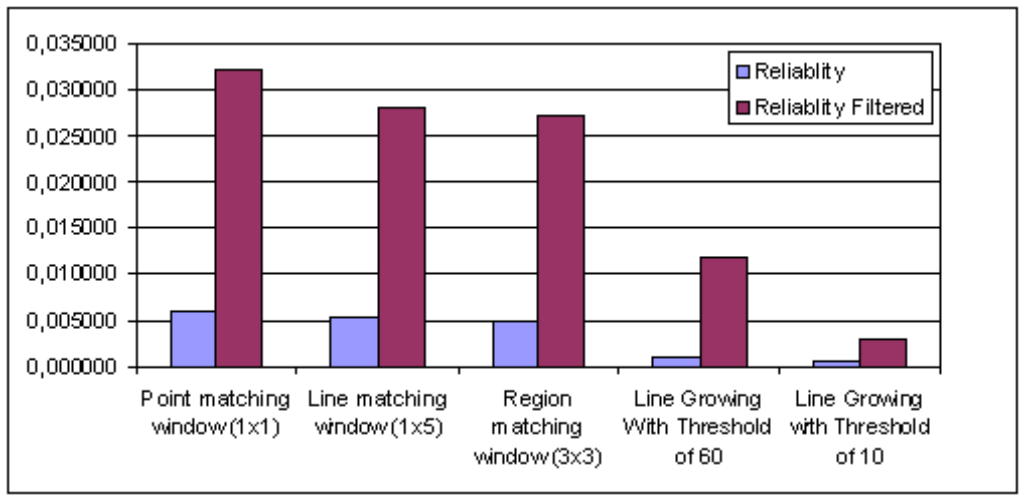

**Figure 12.** Reliability ($R_d$) comparison results illustrated in Figure 5,6,7,8 and 10

Time consumed for calculations of the methods on a personal computer (PC) were compared in the Figure 13.

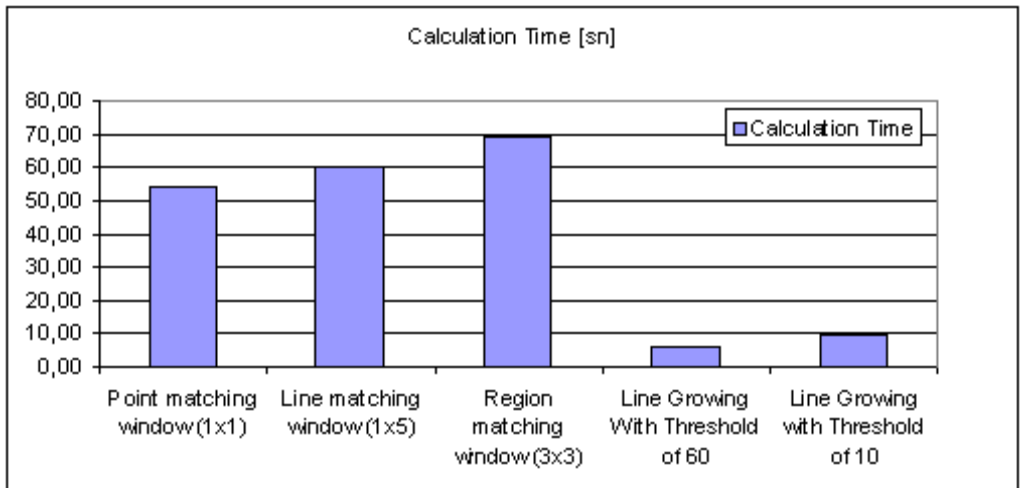

**Figure 13.** Time consumed in calculations of the algorithms by a PC

*Some Remarks*;
- Considering $E_d$ images illustrated in the sub-figures marked (c), error energy is particularly higher at object boundaries, where the disparity is chancing sharply. Because, around the boundaries of the same objects in the stereo pair, there exist regions of which matcher isn't found at the other image of stereo pair. (Occultation) However, these regions reduce the reliability of the stereo matching algorithm, $E_d$ images can be used for detection object boundaries in machine vision applications.
- Method using global error energy minimization by smoothing functions is seen to produce more reliable (See Figure 12) and smoother results. (See Figure 5, 6, 7) But, it is more time consuming for software implementation. Its iterative and simple nature based on repeating sum and squaring operations is rather convenient for hardware implementation (ASIC, FPGA) with distributed computation architecture. Whereas, line growing method is faster for software implementations.
- In the Figure 9 and 11, we present point status of line growing algorithm. Increasing $V_{LG}$ becomes the algorithm tolerant against the error energy and it results increasing line wide in regions and reduce the point in idle status. Therefore, higher $V_{LG}$ makes the disparity estimation smoother without additional cost of computation.
- For the practical application of stereovision in robotic applications, quite smooth disparity estimation would be needed to make the robot vision robust against the faulty decisions in navigation operation. We suggest applying median filtering on the $\tilde{d}$ by a wide windows size in the practical applications. In the Figure 14, 3D view derived from $\tilde{d}$ obtained by global error energy minimization by smoothing functions algorithm. $\tilde{d}$ was smoothed by a median filter with 5x5 window size.

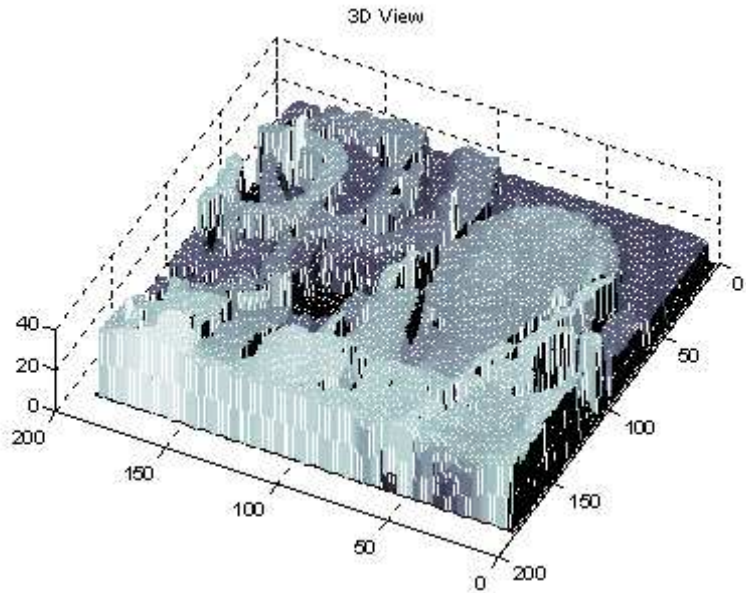

**Figure 14.** 3D view of median filtered $\tilde{d}$ by 5x5 window size

## Conclusions:

We have seen that global error energy minimization by smoothing functions method is more reliable but more time consuming. Better reliability and speed performance was obtained for 1x1 window size (Point matching) in test of error energy minimization by smoothing function methods. Line growing method is more convenient for the sequential computing architectures because of promising higher speed.

Filtering unreliable disparity estimation by average error thresholding was increased reliability of disparity map. Finally, median filtering with large window size makes the disparity and dept maps smoother.

## Propositions:

\* Considering model defined by equations (5), (6), (7), (8) and (9); for a given $\alpha'$ satisfying $\alpha' < \alpha$, it results $R'_d \geq R_d$.

*Proof:*

Lets take an $\alpha'$ satisfying $\alpha' < \alpha$ and for the $\alpha'$, we denote error energy threshold by $V'e$ and denote error energy of the $\tilde{d}'(i,j)$ by $\tilde{E}'_d(i,j)$. According equation (9), we can write following relation,

$$V'e = \alpha' \cdot Mean(E_d) < Ve = \alpha \cdot Mean(E_d)$$

Although, one may find at least one $(i, j)$ point for which $\tilde{E}_d(i,j) > V'e$ and $\tilde{E}'_d(i,j) = ne$, one never find any $(i, j)$ point for which $\tilde{E}'_d(i,j) > Ve$ and $\tilde{E}_d(i,j) = ne$ as a result of elimination according equation (8). Therefore, considering equation (5), we can state that $R'_d \geq R_d$ .

## References:

[1] C. Zitnick and T. Kanade, A Cooperative Algorithm for Stereo Matching and Occlusion Detection, tech. report CMU-RI-TR-99-35, Robotics Institute, Carnegie Mellon University, October, 1999.

[2] H. H. Baker and T. O. Binford, 1981, "Depth from edge and intensity based stereo," In Proc. of the 7th International Joint Conference on Artificial Intelligence, Vancouver, 1981, pp. 631-636.

[3] S. T. Barnard and M. A. Fischler, "Stereo Vision," in Encyclopedia of Artificial Intelligence. New York: John Wiley, 1987, pp. 1083-1090.

[4] D. Scharstein and R. Szeliski. "A taxonomy and evaluation of dense two-frame stereo correspondence algorithms." International Journal of Computer Vision, 47(1/2/3):7-42, April-June 2002.

[5] T.F. Chan, S. Osher, J. Shen, J, "The digital TV filter and nonlinear denoising", Image Processing, IEEE Transaction, Vol.10, pp. 231-241, 2001.

[6] R.C. Gonzalez, R.E. Woods, S.L. Eddins, "Digital Image Processing Second Edition" ,Prentice Hall, pp. 75-142, 2002.

*…to memory of my brother Serdar Onur Alagöz.*